\pdfoutput=1

\documentclass[11pt]{article}
\usepackage{acl}
\usepackage{times}
\usepackage{latexsym}
\usepackage{microtype}

\usepackage{balance}
\usepackage{amsfonts}
\usepackage{amsmath}
\usepackage{amssymb}
\usepackage{amsthm}

\usepackage{graphicx}
\usepackage[utf8]{inputenc} 
\usepackage[T1]{fontenc}    
\usepackage{hyperref}       
\usepackage{url}            
\usepackage{booktabs}       
\usepackage{nicefrac}       
\usepackage{microtype}      
\usepackage{lipsum}
\usepackage{bm}
\usepackage{subfigure}
\usepackage[ruled,vlined]{algorithm2e}
\usepackage{graphics}
\usepackage{epsfig}
\usepackage{adjustbox}
\usepackage{nomencl}
\makenomenclature
\usepackage{etoolbox}
\usepackage{hyperref}
\usepackage{listings}

\usepackage{svg}

\usepackage{multirow}
\usepackage[normalem]{ulem}
\useunder{\uline}{\ul}{}

\usepackage{xargs}                      
\usepackage{cleveref}
\usepackage[colorinlistoftodos,prependcaption,textsize=tiny]{todonotes}

\crefname{section}{§}{§§}
\Crefname{section}{§}{§§}

\newcommand{\eg}{\emph{e.g.}\xspace} 

\def \B {\mathcal{B}}

\def \L {\mathcal{L}}

\def \Q {\mathcal{Q}}

\def \S {\mathcal{S}}

\def \U {\mathcal{U}}

\def \X {\mathcal{X}}
\def \Y {\mathcal{Y}}

\title{Adaptive Ranking-based Sample Selection \\ for Weakly Supervised Class-imbalanced Text Classification}

\author{Linxin Song$^1$, Jieyu Zhang$^2$, Tianxiang Yang$^1$ \and Masayuki Goto$^1$ \\
        $^1$ Waseda University \ \ \ $^2$ University of Washington \\ 
        {\normalsize  songlx.imse.gt@ruri.waseda.jp, \ jieyuz2@cs.washington.edu, \ you\_tensyou@akane.waseda.jp, \ masagoto@waseda.jp}}

\begin{document}
\maketitle

\begin{abstract}
To obtain a large amount of training labels inexpensively,
researchers have recently adopted the weak supervision (WS) paradigm, which leverages labeling rules to synthesize training labels rather than using individual annotations to achieve competitive results for natural language processing (NLP) tasks. 
However, data imbalance is often overlooked in applying the WS paradigm, despite being a common issue in a variety of NLP tasks. 
To address this challenge, we propose Adaptive Ranking-based Sample Selection (ARS2), a model-agnostic framework to alleviate the data imbalance issue in the WS paradigm.
Specifically, it calculates a probabilistic margin score based on the output of the current model to measure and rank the cleanliness of each data point.
Then, the ranked data are sampled based on both class-wise and rule-aware ranking.
In particular, the two sample strategies corresponds to our motivations: (1) to train the model with balanced data batches to reduce the data imbalance issue and (2) to exploit the expertise of each labeling rule for collecting clean samples.
Experiments on four text classification datasets with four different imbalance ratios show that ARS2 outperformed the state-of-the-art imbalanced learning and WS methods, leading to a 2\%-57.8\% improvement on their F1-score.
Our implementation can be found in \url{https://github.com/JieyuZ2/wrench/blob/main/wrench/endmodel/ars2.py}.
\end{abstract}

\section{Introduction}
\begin{figure}[t]
    \centering
    \includegraphics[width=7.7cm]{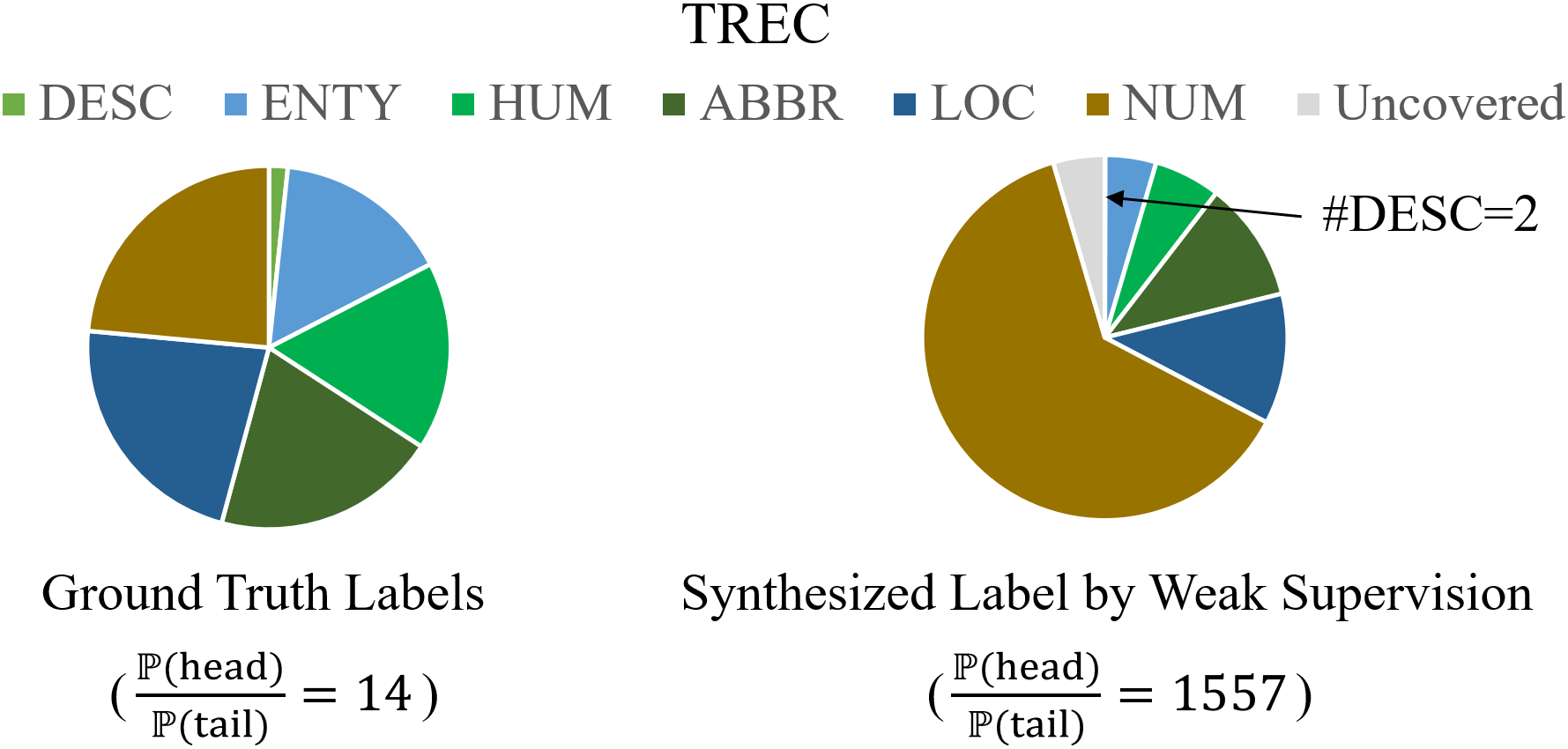}
    \caption{Comparison of class distribution between the ground truth labels and labels produced by weak supervision (WS) on TREC dataset. The uncovered piece represents the data not covered by any labeling rule in WS. It may be observed that WS amplified the class imbalance.}
    \label{fig:intro}
\end{figure}
Deep learning models rely heavily on high-quality yet expensive, labeled data. 
Owing to this considerable cost, the weak supervision (WS) paradigm has increasingly been used to reduce human efforts~\cite{ratner2016data, zhang2021wrench}. This approach synthesizes training labels with labeling rules to significantly improve the efficiency of creating training sets and have achieved competitive results in natural language processing (NLP)~\cite{cosine, denoise, ruhling2021end}.
However, existing methods leveraging the WS paradigm to perform NLP tasks mostly focus on reducing the noise in training labels brought by labeling rules, while ignoring the common and critical problem of data imbalance. 
In fact, in a preliminary experiment performed as part of the present work (Fig.~\ref{fig:intro}), we found that the WS paradigm may amplify the imbalance ratio of the dataset because the synthesized training labels tend to have more imbalanced distribution.

To address this issue, we propose ARS2 as a general model-agnostic framework based on the WS paradigm. 
ARS2 is mainly divided in two steps, including (1) warm-up, in which stage noisy data is used to train the model and obtain a noise detector; (2) continual training with adaptive ranking-based sample selection. 
In this stage, we use the noise detector trained in the warm-up stage to evaluate the cleanliness of the data, and use the ranking obtained based on this evaluation to sample the data.
We followed previous works \citet{ratner2016data, denoise, zhang2021creating} in using heuristic programmatic rules to annotate the data.
In weak supervised learning, researchers use a label model to aggregate weak labels annotated by rules to estimate the probabilistic class distribution of each data point. 
In this work, we use a label model to integrate the weak labels given by the rules as pseudo-labels during the training process to obviate the need for manual labeling. 

To select the samples most likely to be clean, we adopt a selection strategy based on small-loss, which is a very common method that has been verified to be effective in many situations~\cite{jiang2018mentornet, yu2019does, yao2020searching}. Specifically, deep neural networks, have strong ability of memorization~\cite{wu2018light, wei2021crest}, will first memorize labels of clean data and then those of noisy data with the assumption that the clean data are of the majority in a noisy dataset. 
Data with small loss can thus be regarded as clean examples with high probability. 
Inspired by this approach, we propose probabilistic margin score (PMS) as a criterion to judge whether data are clean. 
Instead of using the confidence given by a model directly, a confidence margin is used for better performance~\cite{ye2020identifying}. We also  performed a comparative experiment on the use of margin versus the direct use of confidence, as described in Sec.~\ref{sec:mainres}.

Sample selection based on weak labels can lead to severe class imbalance. 
Consequently, models trained using these imbalanced subsets can exhibit both superior performance on majority classes and inferior performance on minority classes~\cite{effective_number}. A reweighted loss function can partially address this problem. 
However, performance remains nonetheless limited by noisy labels, that is, data with majority-class features may be annotated as minority-class data incorrectly, which misleads the training process. 
Therefore, we propose a sample selection strategy based on class-wise ranking (CR) to address imbalanced data.
Using this strategy, we can select relatively balanced sample batches for training and avoid the strong influence of the majority class.

To further exploit the expertise of labeling rules, we also propose another sample selection strategy called rule-aware ranking (RR). 
We use aggregated labels as pseudo-labels in the WS paradigm and discards weak labels. 
However, the annotations generated by rules are likely to contain a considerable amount of valid information. 
For example, some rules yield a high proportion of correct results. 
The higher the PMS, the more likely the labeling result of the rules is to be close to the ground truth. 
Using this strategy, we can select batches with clean data for training and avoid the influence of noise.

The primary contributions of this work are summarized as follows. (1) We propose a general, model-agnostic weakly supervised leading framework called ARS2 for imbalanced datasets; (2) we also propose two reliable adaptive sampling strategies to address data imbalance issues. (3) The results of experiments on four benchmark datasets are presented to demonstrate that the ARS2 improved on the performance of existing imbalanced learning and weakly supervised learning methods, by 2\%-57.8\% in terms of F1-score.
\section{Weakly Supervised Class-imbalanced Text Classification}
\subsection{Problem Formulation}
In this work, we study class-imbalanced text classification in a setting with weak supervision. Specifically, we consider an unlabeled dataset $\mathcal{D}$ consisting of $N$ documents, each of which is denoted by $x_i\in \X$. For each document $x_i$, the corresponding label $y_i\in\mathcal{Y}=\{1,2,...,C\}$ is unknown to us, whereas the class prior $p(y)$ is given and highly imbalanced. Our goal is to learn a parameterized function $f(\cdot; \bm \theta):\X \xrightarrow{} \Delta^{C}$\footnote{$\Delta^{C}$ is a $C$-dimension simplex.} which outputs the class probability $p(y\mid x)$ and can be used to classify documents during inference.

To address the lack of ground truth training labels, we adopt the two-stage weak supervision paradigm~\cite{Ratner16, zhang2021wrench}. In particular, we rely on $k$ user-provided heuristic rules $\{r_i\}_{i\in\{1,...,k\}}$ to provide weak labels.
Each rule $r_i$ is associated with a particular label $y_{r_i}\in\mathcal{Y}$, and we denote by $l_i$ the output of the rule $r_i$. It either assigns the associated label ($l_i=y_{r_i}$) to a given document or abstains ($l_i=-1$) on this example.
Note that the user-provided rules could be noisy and conflict with one another. 
For the document $x$, we concatenate the output weak labels of $k$ rules $l_1,...,l_k$ as $\bm{l}_x$.
Throughout this work, we apply the weak labels output by heuristic rules to train a text classifier.

\begin{figure*}[!t]
    \centering
    \includegraphics[width=\textwidth]{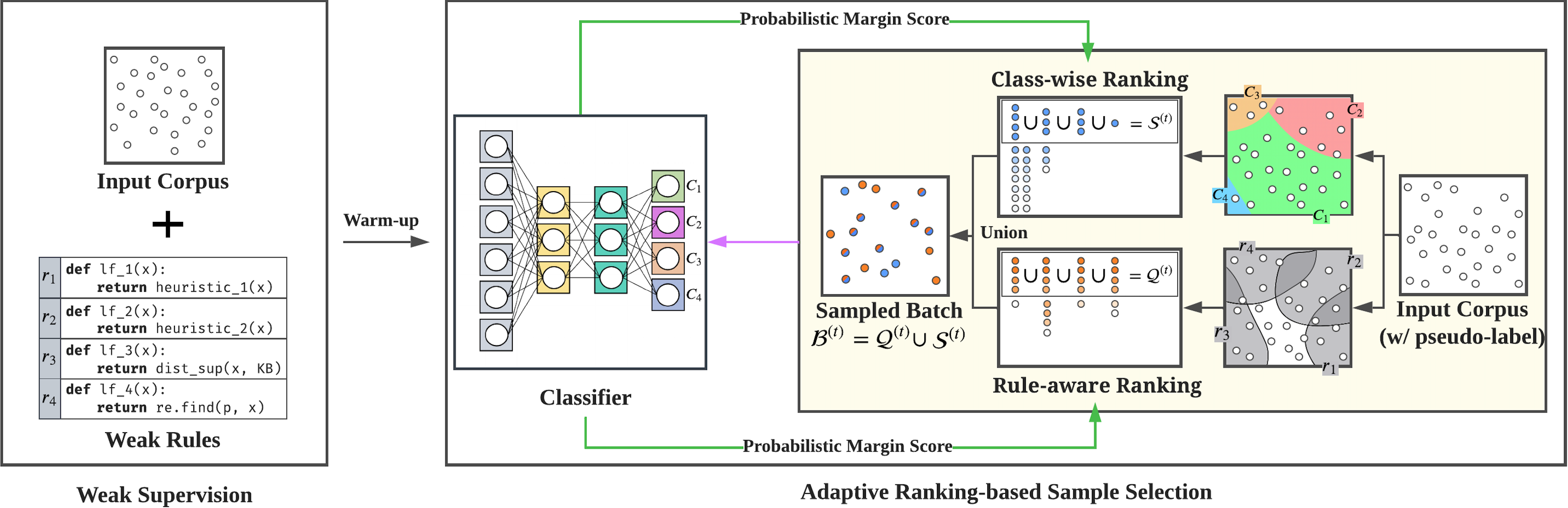}
    \caption{Overview of ARS2. Our framework has two stages: (1) warm-up, which is used to let the model learn how to distinguish noisy data; (2) continual training with adaptive sampling, which is used to sample clean data. We adopt two different adaptive sampling strategies, including class-wise ranking sampling and rule-aware ranking sampling.}
    \label{fig:main}
\end{figure*}

\subsection{Aggregation of Weak Labels}
Label models are used to aggregate weak labels under the weak supervision paradigm, which are in turn used to train the desired end model in the next stage. 
Existing label models include Majority Voting (MV), Probabilistic Graphical Models (PGM)~\cite{dawid1979maximum, ratner2019training, fu2020fast}, etc.
In this research, we use PGM implement by~\citet{ratner2019training} as our label model $g(\cdot)$, which can be described as
\begin{equation}
    g(\bm{l}_x) = \mathbb{P}(y \mid \bm{l}_x).
    \label{eq:lm}
\end{equation}
This assumes that $\bm{l}_x$ as a random variable for label model. After modeling the relationship between the observed variable $\bm{l}_x$ and unobserved variable $y$ by Bayes methods, the label model obtains the posterior distribution of $y$ given $\bm{l}_x$ by inference process like expectation maximization or variational inference. Finally, we set the maximum value of $\mathbb{P}(y \mid \bm{l}_x)$ as the hard pseudo-label $\Tilde{y}_x$ of $x$ for later end model training.

\subsection{Adaptive Ranking-based Sample Selection}
We propose an adaptive ranking sample selection approach to simultaneously solve the problems caused by data imbalance and those resulting from the noise generated by the application of procedural rules.
First, the classification model $f(\cdot; \bm \theta)$ is warmed up with pseudo-labels $\Tilde{y}_x$, which are used to train the model as a noise indicator that can discriminate noise in the next stage. 
Then, we continue training the warmed-up model by using the data sampled by adaptive data sampling strategies, including class-wise ranking (CR) and rule-aware ranking (RR) supported by probabilistic margin score (PMS). 
The training procedures are summarized in Algorithm \ref{alg:main} and the proposed framework is illustrated in Figure~\ref{fig:main}.

\begin{algorithm}[t]
	\begin{small}
	\KwIn{Weak labeled training data $\X$; Pseudo label $\Tilde{y}$; Classification model $f(\cdot; \bm \theta)$.}
	
	// \textit{Warm-up $f(\cdot, \bm\theta)$ with weak labeled data.} \\
	\For{$t = 1, 2, \cdots, T$}{
    	1. Sample a minibatch $\B$ from $\X$.\\
    	2. Update $\theta$ by Eq.~(\ref{eq:warmup}) before early stop.
    }
    
	// \textit{Continue training with sample selection.} \\
	\For{$t = 1, 2, \cdots, T_s$}{
    	1. Calculate score for all $x\in\X$ (Sec.~\ref{sec:pms}). \\
    	2. Sample $\Q^{(t)}$ from $\X$ (Sec.~\ref{sec:cwr}). \\
    	3. Sample $\S^{(t)}$ from $\X$ (Sec.~\ref{sec:rcc}). \\
    	4. Update $\theta$ using $\U^{(t)}=\Q^{(t)}\cup\S^{(t)}$ by Eq.~(\ref{eq:finetune}).
    }
	\KwOut{Output final model $f(\cdot; \bm \theta)$.}
	\end{small}
	\caption{ARS2}
	\label{alg:main}
\end{algorithm}
\paragraph{Warm-up.}
Inspired by \citet{zheng2020error}, the prediction of a noisy classification model can be a good indicator of whether the label of a training data point is clean. 
Our method relies on a noise indicator trained at warm-up to determine whether each training data is clean.
However, a model with sufficient capacity (e.g., more parameters than training examples) can “memorize” each example, overfitting the training set and yielding poor generalization to validation and test sets~\cite{goodfellow2016deep}.
To prevent a model from overfitting noisy data, we warm-up the model $f(\cdot; \bm \theta)$ with early stopping~\cite{dodge2020fine}, and solve the optimization problem by
\begin{equation}
    \min_{\bm\theta} \frac{1}{N} \sum_{x \in \X} \L(f(x; \bm\theta), \Tilde{y}_x),
    \label{eq:warmup}
\end{equation}
where $\L$ denotes a loss function and $\Tilde{y}_x$ is pseudo-label aggregated by a label model. In this research, we do not limit the definition of the loss function; that is, any loss function suitable for a multi-classification task can be used. 

\paragraph{Continual training with sample selection.}
Noisy and imbalanced labels impair the predictive performance of learning models, especially deep neural networks, which tend to exhibit a strong memorization capability~\cite{wu2018light, wei2021crest}.
To reduce the influence of imbalance and noise in the classification model, we adopt the intuitive approach of simply continuing the training process with clean and balanced data.
In the continual training phase, we filter for clean data filter using a warmed-up model and sample data in a balanced way to calibrate the model. This procedure provides a more robust performance on a noisy, imbalanced dataset.
To achieve this, we propose a measurement of label quality called probabilistic margin score (PMS) (Sec.~\ref{sec:pms}), using two sample selection strategies, including class-wise ranking (CR) (Sec.~\ref{sec:cwr}) and rule-aware ranking (RR) (Sec.~\ref{sec:rcc}).
We adopt batch sampling by using CR as $\S$ and RR as $\Q$ after several steps, using batch $\U = \Q\cup\S$ to continue training the model by solving the following optimization problem.
\begin{equation}
    \min_{\bm\theta} \frac{1}{|\U|} \sum_{x \in \U} \L(f(x; \bm\theta),
    \Tilde{y}_x).
    \label{eq:finetune}
\end{equation}

\paragraph{Dynamic-Rate Sampling}
During the training procedure, the model learns from data with easy/frequent patterns and those with harder/irregular patterns in separate training stages~\cite{arpit2017closer}.
According to \citet{chen2020self}, a small amount of clean data should be selected first, and then a larger amount of clean data can be selected after the model has been trained well.
Therefore, instead of fixing the number of selected data, as the continual training stage proceeds, we linearly increase the size of the training batch $\U$ by
\begin{equation}
    k = \frac{B}{C} \times (1 + \frac{t(\gamma - 1)}{T_s}), 
\end{equation}
where $B$ indicates batch size. The sampling ratio $\gamma$ ranged from 1 to 10 in our experiments.

\subsection{Probabilistic Margin Score}
\label{sec:pms}
To measure the impact of the training label on label quality, inspired by~\citet{pleiss2020identifying}, we use the margin of output predictions to identify data with low label quality for later adaptive ranking-based sample selection.
The predictions of the network can be regarded as a measurement of the quality of labels, which means that a correct prediction indicates high label quality, and vice versa~\cite{jiang2018mentornet, yu2019does, yao2020searching}.
Moreover, margins are advantageous because they are simple and efficient to compute during training, and they naturally factorize across samples, which makes it possible to estimate the contribution of each data point to label quality.

Base on this idea, we propose the probabilistic margin score (PMS) to reflect the quality of a given label. PMS is formulated as
\begin{equation}
    s(x) = f_{\Tilde{y}}(x;\bm \theta) - \underset{y\in \Y \setminus\{\Tilde{y}\}}{\text{max}} f_y(x; \bm \theta),
    \label{eq:score}
\end{equation}
where $f_y(\cdot; \bm \theta)$ denotes the prediction of $y$ by classification model $f(\cdot; \bm\theta)$ and $s(x)\in[-1, 1]$. A negative margin corresponds to an incorrect prediction, and vice versa. The margin captures how much lager the (potentially incorrect) assigned confidence is than all other confidences, which indicates the label quality of $x$.

\subsection{Class-wise Ranking}
\label{sec:cwr}
If the top-ranked data are directly selected based on PMS to form the training batch, the majority-class data are highly like to dominate the selected batch. This class imbalance in the data results in suboptimal performance. 
To overcome this problem, we propose class-wise ranking (CR). Specifically, instead of naively selecting the top-k data ranked by PMS from the entire training set, we select top-ranked data from each class to maintain the class balance of the resultant batch. 
However, CR might introduce more noise if we attempted to construct a perfectly class-balanced data batch. Thus, to reduce the noise while ensuring class balance as much as possible, we set a threshold on PMS to filter the noisy data.

We assume that the set of data belonging to class $y_i$ is $\X_{y_i}$ and extract the top-k ranked data from each $\X_{y_i}$ as $\S^{(t)}_{y_i}$ at step $t$ according to the ranking result from $s(\X)$. We also set a threshold $\xi$ for $s(\X)$ to filter noisy data.
\begin{equation}
    \S^{(t)}_{y_i} = \underset{x\in \X_{y_i}, s(x) > \xi}{\text{top-k}} \ s(x).
    \label{eq:cr1}
\end{equation}
Then, we concatenate $\S^{(t)}_{y_i}$ drawn from each class as a new batch $\S^{(t)}$ as given below.
\begin{equation}
    \S^{(t)} = \bigcup_{y_i\in \Y} \S^{(t)}_{y_i}.
    \label{eq:cr2}
\end{equation}

\subsection{Rule-aware Ranking}
\label{sec:rcc}
As discussed above, the training labels are synthesized from multiple labeling rules. In the proposed framework, these labeling rules are used not only to produce training labels as in a typical WS pipeline, but also to guide the sample selection.
Specifically, each labeling rule typically assigns a certain label, \eg, \emph{sports}, to only a part of the training set based on its expertise. Within this part of the dataset, some of the data must actually belong to the class \emph{sports}; otherwise, the relevant labeling rule would be useless.
Thus, we propose rule-aware ranking (RR) to separately rank the data covered by each labeling rule and then extract the top-ranked data from each individual ranking. We plan to elaborate on this selection process in a sequel to the present work.

We denote the set of data covered by each labeling rule $r_i$ as $\X_{r_i}$. We then rank $\X_{r_i}$ based on PMS $s(x)$ and extract the top-k data from each $\X_{r_i}$ as $\Q^{(t)}_{r_i}$ at step $t$. 
\begin{equation}
    \Q^{(t)}_{r_i} = \underset{x\in \X_{r_i}}{\text{top-k}} \ s(x).
    \label{eq:rr1}
\end{equation}
Specifically, to avoid the error-propagation, we use the weak label $l_x$ instead of $\Tilde{y}_x$ as the new label of $x\in\Q^{(t)}_{r_i}$ when $l_x$ is unipolar.
We claim that data with high confidence will also have a high probability of weak labels being equal to ground truth labels, especially when multiple rules assign the same weak label to the data.
The main reason is that those weak labels are not influenced by the training process and are weakly supervised by human level.
Then, we take the union of $\Q^{(t)}_{r_i}$ drawn from each subset as a new batch $\Q^{(t)}$.
\begin{equation}
    \Q^{(t)} = \bigcup_{r_i\in \bm{r}} \Q^{(t)}_{r_i}.
    \label{eq:rr2}
\end{equation}

\section{Experiment}
\subsection{Experiment Setup}
\paragraph{Tasks and Datasets.}
To evaluate the proposed methods, we used four open benchmark datasets, including AGNews (news topic classification), Yelp (sentiment classification)~\cite{agnewsyelp}, TREC (question classification~\citet{li2002learning}) and ChemProt (relation classification~\citet{krallinger2017overview}). 
Specifically, each dataset was weakly annotated by several rules provided by~\citet{denoise, awasthi2020learning, cosine}. 
The relevant statistic for each dataset are shown in Table~\ref{tab:dataset}.
\begin{table}[!htb]
    \centering
\begin{adjustbox}{max width=0.48\textwidth}
\begin{tabular}{c|c|c|c|c|c|c}
\toprule
    Dataset & Task & \# Class & \# Rule & \# Train & \# Valid & \# Test  \\\midrule
    AG News & News Topic Class. & 4 & 9 & 96k & 12k & 12k \\
    Yelp & Sentiment Class. & 2 & 8 & 30.4k & 3.8k & 3.8k \\
    TREC & Question Class. & 6 & 68 & 4.9k & 500 & 500 \\
    ChemProt & Relation Class. & 10 & 26 & 12.8k & 1.6k & 1.6k \\
    \bottomrule
\end{tabular}
\end{adjustbox}
\caption{Dataset Statistics.}
\label{tab:dataset}
\end{table}

\paragraph{Imbalance Learning Setups.}
Following \citet{effective_number}, we created an imbalanced version of AGNews and Yelp by reducing the training and validation examples for each class with an exponential function according to the ground truth data labels. 
We set four different imbalance ratios $\rho=\max_y \mathbb{P}(x) / \min_y \mathbb{P}(x)$ for AGNews and Yelp in 1, 10, 20 and 50, respectively. 
We used the original versions of TREC and Chemprot, as they are known to be imbalanced. 
Relevant statistics on the imbalanced datasets are shown in \ref{appendix:data}.

\paragraph{Weak Supervision Setups.}
In weak supervision, assume that the data is not artificially annotated, the labels are annotated by a label model instead of using the ground truth labels. 
The label model can analyze the results of rule-based annotation and output the most likely label for a given data sample. 
Throughout the experiments, we used Snorkel \cite{ratner2019training} as the label model to aggregate the outputs of labeling rules.

\begin{table*}[t]
    \centering
    \scalebox{0.53}{
    \hspace{-3mm}
    \renewcommand{\arraystretch}{1}
    \begin{tabular}{ c | c c c c | c c c c | c | c }
    \toprule
         \multicolumn{1}{c}{} &
         \multicolumn{4}{c|}{\textbf{Agnews (Imbalance Ratio ($\downarrow$))}} &
         \multicolumn{4}{c|}{\textbf{Yelp (Imbalance Ratio ($\downarrow$))}} &
         \multicolumn{1}{c|}{\textbf{TREC}} &
         \multicolumn{1}{c}{\textbf{Chemprot}}\\
    \midrule
    \multirow{1}{0pt}{}
        \textbf{Method ($\downarrow$)} & 1 & 10 & 20 & 50 & 1 & 10 & 20 & 50 & - & - \\
    \midrule
    \multirow{10}{0pt}{}
        {CE+LA} & $0.854\pm0.001$ & $0.843\pm0.005$ & $0.826\pm0.010$ & $0.753\pm0.037$ & $0.914\pm0.003$ & $0.877\pm0.016$ & $0.694\pm0.097$ & $0.515\pm0.145$ & $0.394\pm0.033$ & $0.340\pm0.004$ \\
        {CE+EN} & $0.857\pm0.003$ & $0.834\pm0.007$ & $0.824\pm0.005$ & $0.772\pm0.016$ & $0.916\pm0.002$ & $0.723\pm0.200$ & $0.806\pm0.012$ & $0.658\pm0.036$ & $0.450\pm0.005$ & $0.389\pm0.030$ \\
        {CE+EN+LA} & $0.854\pm0.007$ & $0.836\pm0.005$ & $0.838\pm0.004$ & $0.786\pm0.020$ & $0.919\pm0.002$ & $0.610\pm0.204$ & $0.807\pm0.013$ & $0.699\pm0.023$ & $0.409\pm0.152$ & $0.336\pm0.010$ \\
        {Dice} & $0.855\pm0.003$ & $0.846\pm0.006$ & $0.746\pm0.003$ & $0.590\pm0.033$ & $0.882\pm0.003$ & $0.821\pm0.048$ & $0.637\pm0.241$ & $0.323\pm0.000$ & $0.476\pm0.022$ & $0.233\pm0.010$ \\
        {LDAM} & $0.804\pm0.090$ & $0.828\pm0.012$ & $0.813\pm0.009$ & $0.100\pm0.002$ & $0.819\pm0.052$ & $0.751\pm0.042$ & $0.388\pm0.100$ & $0.342\pm0.002$ & $0.154\pm0.033$ & $0.356\pm0.005$ \\
        \midrule
        {COSINE} & 0.854 $\pm$ 0.004 & 0.720 $\pm$ 0.068 & 0.822 $\pm$ 0.003 & 0.574 $\pm$ 0.004 & 0.912 $\pm$ 0.001 & 0.496 $\pm$ 0.192 & 0.836 $\pm$ 0.008 & 0.820 $\pm$ 0.000 & 0.477 $\pm$ 0.001 & 0.346 $\pm$ 0.009 \\
        {Denoise} & 0.852 $\pm$ 0.001 & 0.537 $\pm$ 0.111 & 0.526 $\pm$ 0.112 & 0.471 $\pm$ 0.188 & 0.811 $\pm$ 0.004 & 0.343 $\pm$ 0.004 & 0.498 $\pm$ 0.126 & 0.323 $\pm$ 0.000 & 0.236 $\pm$ 0.021 & 0.259 $\pm$ 0.075 \\
        \midrule
        \midrule
        {ARS2 (w/o CR\&RR)} & 0.854 $\pm$ 0.001 & 0.840 $\pm$ 0.006 & 0.818 $\pm$ 0.035 & 0.810 $\pm$ 0.024 & 0.917 $\pm$ 0.006 & 0.868 $\pm$ 0.011 & 0.822 $\pm$ 0.057 & 0.738 $\pm$ 0.066 & 0.342 $\pm$ 0.033 & 0.396 $\pm$ 0.021 \\
        {ARS2 (w/o RR)} & 0.854 $\pm$ 0.000 & 0.840 $\pm$ 0.005 & 0.797 $\pm$ 0.078 & 0.777 $\pm$ 0.023 & 0.920 $\pm$ 0.003 & 0.844 $\pm$ 0.021 & 0.764 $\pm$ 0.096 & 0.794 $\pm$ 0.054 & 0.348 $\pm$ 0.028 & 0.402 $\pm$ 0.033 \\
        {ARS2 (w/o CR)} & \textbf{0.884 $\pm$ 0.006} & 0.842 $\pm$ 0.005 & 0.818 $\pm$ 0.022 & 0.759 $\pm$ 0.069 & 0.929 $\pm$ 0.002 & 0.861 $\pm$ 0.031 & 0.750 $\pm$ 0.132 & 0.555 $\pm$ 0.165 & 0.490 $\pm$ 0.065 & 0.303 $\pm$ 0.004 \\
        {ARS2 (Conf.)} & 0.854 $\pm$ 0.001 & 0.847 $\pm$ 0.006 & 0.821 $\pm$ 0.015 & 0.819 $\pm$ 0.018 & \textbf{0.939 $\pm$ 0.001} & 0.908 $\pm$ 0.004 & 0.851 $\pm$ 0.028 & 0.835 $\pm$ 0.034 & 0.574 $\pm$ 0.027 & 0.342 $\pm$ 0.003 \\
        {ARS2} & 0.882 $\pm$ 0.003 & \textbf{0.859 $\pm$ 0.012} & \textbf{0.844 $\pm$ 0.035} & \textbf{0.827 $\pm$ 0.011}  & 0.936 $\pm$ 0.002 & \textbf{0.910 $\pm$ 0.003} & \textbf{0.852 $\pm$ 0.024} & \textbf{0.854 $\pm$ 0.012} & \textbf{0.597 $\pm$ 0.025} & \textbf{0.404 $\pm$ 0.006} \\
    \bottomrule
    \end{tabular}
    }
    \caption{\textbf{F1-macro result on 2-layer MLP}. Comparison among imbalance learning methods, weak supervision methods, and ASR2 (as well as its variants). The ranking of using confidence is identical to that ranked by negative loss. Note that ARS2 outperform all baselines in all imbalanced datasets.}
    \label{tab:mlpres}
\end{table*}

\paragraph{Implementation Details.}
We choose the Multi-Layer Perceptron (MLP) with 2 hidden layers and RoBERTa~\cite{liu2019roberta} as the backbone language model for our method in all baselines. In the case of using RoBERTa as a backbone model, rather than training the RoBERTa as a noise indicator in warm-up stage, we used a warmed-up MLP as a noisy indicator to sample training batches for RoBERTa because a large model may easily overfit noisy data in a few steps. We used the classification macro-average F1 score on the test set as the evaluation metric for all datasets. We implemented our method using PyTorch~\cite{paszke2019pytorch} with the WRENCH code-base~\cite{zhang2021wrench}\footnote{Our implementations will be released upon the acceptance of this work.}.

\subsection{Baselines}
\paragraph{Imbalance Learning Methods:} 
(1) \textbf{Logit Adjustment} (LA)~\cite{logit_adjustment}: This method combines post-hoc weight normalization and loss modification to balance head and tail classes by adding a class-wise offset to the loss. 
(2) \textbf{Effective Number} (EN)~\cite{effective_number}: This method uses the proportion of sampled data as the class-wise weight of loss function, which helps the model to learn a useful decision boundary. 
(3) \textbf{Dice Loss} (Dice)~\cite{li2019dice}: The dice loss method uses dynamically adjusted weight to improve the S$\phi$rensen-Dice coefficient (a method that values both false positive and false negative and works well for imbalanced datasets), which reduces the impact of easy negative data during training. 
(4) \textbf{LDAM}~\cite{ldam}: This method encourages the model
to treat the optimal trade-off problem between per-class margins. 
(5) \textbf{Effective Number + Logit Adjustment}: Because the effective number is a class-wise re-weighting method, we have also combined this method with logit adjustment as a baseline.

\paragraph{Weak Supervision Methods:} 
(1) \textbf{COSINE}~\cite{cosine} The COSINE method uses weakly labeled data to fine-tune pre-trained language models by contrastive self-training. 
(2) \textbf{Denoise}~\cite{denoise}: This method estimates the source reliability
using a conditional soft attention mechanism and then reducing label noise by aggregating weak labels annotated by a set of rules.

\subsection{Main Result}
\label{sec:mainres}
Our main results on a 2-layer MLP model are reported in Table \ref{tab:mlpres}. Our method outperformed all the baselines on both balanced and imbalanced datasets. 
The results of the experiment also indicated the following. 

The performance of all baseline methods generally showed a downward trend with increasing imbalance ratio, and the decline was more evident in the binary classification problem of Yelp. 
In contrast, on the multi-classification problem dataset AGNews, although the head class and the tail class exhibited a relatively large gap because the number of each class declined step-by-step, the ratio between the head class and the second head class did not reach the value of the imbalance ratio. 
Therefore, the model's performance decline on multi-classification was slightly smaller than in binary classification.

\begin{figure*}[!t]
    \centering
    \subfigure[$\rho=10$]{
        \includegraphics[width=5.05cm]{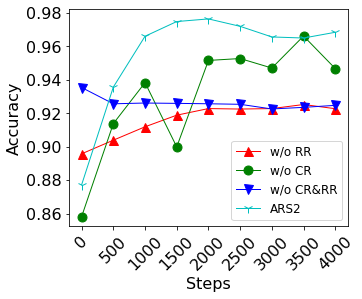}
    }
    \subfigure[$\rho=20$]{
        \includegraphics[width=4.8cm]{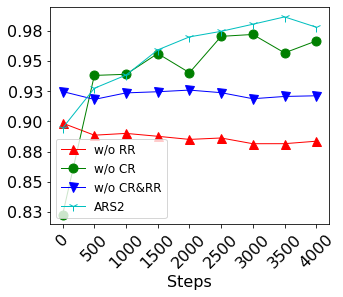}
    }
    \subfigure[$\rho=50$]{
        \includegraphics[width=4.8cm]{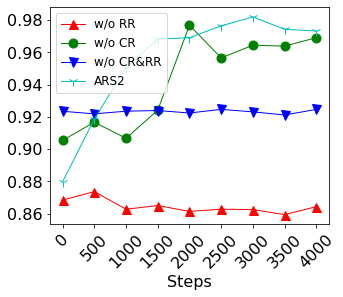}
    }
    \caption{The quality of the selected data in \textbf{AGNews} under each imbalance ratio.}
    \label{fig:agnewsabla}
\end{figure*}

\begin{figure*}[!t]
    \centering
    \subfigure[ $\rho=10$]{
        \includegraphics[width=5.05cm]{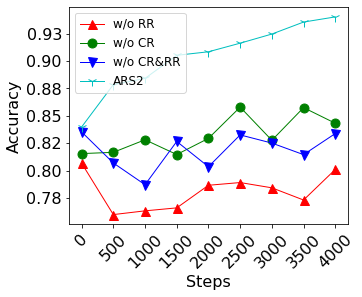}
    }
    \subfigure[ $\rho=20$]{
        \includegraphics[width=4.8cm]{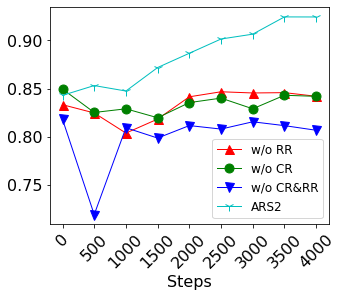}
    }
    \subfigure[ $\rho=50$]{
        \includegraphics[width=4.8cm]{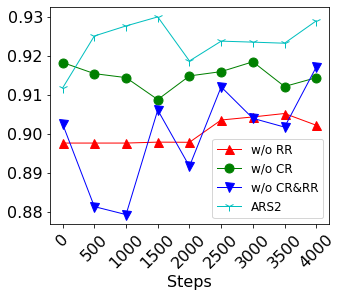}
    }
    \caption{The quality of the selected data in \textbf{Yelp} under each imbalance ratio.}
    \label{fig:yelpabla}
\end{figure*}

\begin{figure}[t]
    \centering
    \subfigure[Linear Ratio]{
        \includegraphics[width=3.7cm]{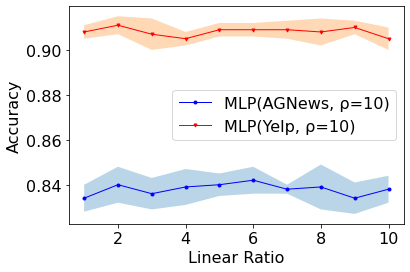}
        \label{fig:param_lr}
    }\hspace{-2.3mm}
    \subfigure[Threshold]{
        \includegraphics[width=3.55cm]{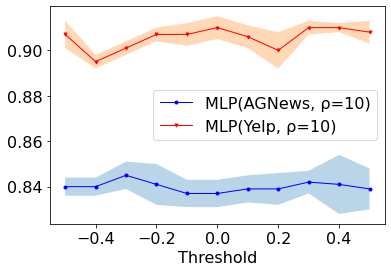}
        \label{fig:param_thres}
    }
    \caption{Hyperparameter Study}
\end{figure}

On the balanced dataset, the performance of ARS2 was very similar to that of using RR only, whereas the performance of ARS2 (w/o CR) was slightly better than that of ARS2 in AGNews. 
This occurred because CR did not operate as intended on the balanced dataset but instead affected the diversity of data, resulting in a small gap between ARS2 and RR on the balanced set. 
The role of CR became gradually more prominent with increasing imbalance ratio because CR was able to balance the original unbalanced sample set such that the model was not affected by the data from the majority class, improving performance.

The baseline performance gradually weakened with the increase in the imbalance ratio. These baselines were all loss-modification-related methods. In the noisy dataset, if a data point originally belonging to the head class is incorrectly marked as belonging to a tail class, the loss modification feature causes the model to assign a larger weight to the data that may have noise, which deepens the error of the model, and results in poor training performance.

\subsection{Ablation Study}
\label{sec:abla}
We also examined the importance of CR, RR, and different measurements of label quality. The 8th, 9th and 11th rows of Table~\ref{tab:mlpres} summarize the results. 
It may be observed that the performance of ARS2 without CR on the balanced set was almost the same as that of ARS2, which shows that RR was the main contributor to the balanced set. 
However, the contributions of both start to converge with increasing imbalance ratio. 
This shows that the data filtered by CR can allow the model to learn more balance features from the imbalance dataset, whereas RR can make the process smoother and reduce the model's misjudgment of high-ranking data. 

We also compared the performance of two ranking scores. The 11th and 12th rows of Table~\ref{tab:mlpres} show that ARS2 with PMS performed better than with confidence, except for the Yelp dataset with an imbalance ratio of $1$. However, the difference in performance in this situation was less than 1\%. This shows that PMS can reflect the cleanliness of data better, and thus may be considered a more useful ranking score.

\subsection{Hyperparameter Study}
We studied two hyperparameters, namely $\gamma$ and $\xi$ used in controlling the quality of adaptive sampling. 
We set the linear ratio from $1$ to $10$, where higher values of the linear ratio indicate larger amounts of data in the sampled batch. 
We also set the threshold from -0.5 to 0.5, which indicated the quality of the data of our sampled batch. 
Figures~\ref{fig:param_lr} shows that ARS2 is insensitive to $\gamma$ as the performance is less than 0.8\%. In the case of Yelp, ARS2 cannot perform well with a small threshold (less than 0) because the loose threshold may lead a large amount of noisy data to affect the training process. While in the case of AGNews, the high quality of rules helps RR to re-corrected the noisy label, increase the clean data points accessed by model during training.

\begin{table*}[t]
    \centering
    \scalebox{0.55}{
    \renewcommand{\arraystretch}{1} 
    \begin{tabular}{ c c c c c | c c c c | c | c }
    \toprule
         \multicolumn{1}{c}{} &
         \multicolumn{4}{c|}{\textbf{AGNews (Imbalance Ratio ($\downarrow$))}} &
         \multicolumn{4}{c|}{\textbf{Yelp (Imbalance Ratio ($\downarrow$))}} & 
         \textbf{TREC} &
         \textbf{Chemprot} \\
    \midrule
    \multirow{1}{0pt}{}
        \textbf{Method ($\downarrow$)} & 1 & 10 & 20 & 50 & 1 & 10 & 20 & 50 & - & -\\
    \midrule
    \multirow{10}{0pt}{}
        {CE+LA}     & 0.858 $\pm$ 0.002 & 0.847 $\pm$ 0.004 & 0.833 $\pm$ 0.009 & 0.823 $\pm$ 0.030 &   0.949 $\pm$ 0.005 & 0.900 $\pm$ 0.014 & 0.837 $\pm$ 0.103 & 0.323 $\pm$ 0.000 & 0.545 $\pm$ 0.014 & 0.349 $\pm$ 0.019  \\
        {CE+EN}     & 0.859 $\pm$ 0.002 & 0.841 $\pm$ 0.005 & 0.852 $\pm$ 0.006 & 0.823 $\pm$ 0.020 &   0.941 $\pm$ 0.014 & 0.897 $\pm$ 0.026 & 0.653 $\pm$ 0.270 & 0.323 $\pm$ 0.000 & 0.561 $\pm$ 0.022 & 0.453 $\pm$ 0.004  \\
        {CE+EN+LA}  & 0.861 $\pm$ 0.001 & 0.851 $\pm$ 0.004 & 0.838 $\pm$ 0.007 & 0.809 $\pm$ 0.051 &   0.947 $\pm$ 0.002 & 0.882 $\pm$ 0.058 & 0.839 $\pm$ 0.076 & 0.323 $\pm$ 0.000 & 0.505 $\pm$ 0.068 & 0.039 $\pm$ 0.018  \\
        {Dice}      & 0.874 $\pm$ 0.001 & 0.875 $\pm$ 0.002 & 0.865 $\pm$ 0.005 & 0.800 $\pm$ 0.064 &   0.939 $\pm$ 0.007 & 0.628 $\pm$ 0.267 & 0.848 $\pm$ 0.102 & 0.323 $\pm$ 0.000 & 0.455 $\pm$ 0.029 & 0.064 $\pm$ 0.008  \\
        {LDAM}      & 0.859 $\pm$ 0.002 & 0.823 $\pm$ 0.007 & 0.815 $\pm$ 0.004 & 0.836 $\pm$ 0.009 &   0.943 $\pm$ 0.008 & 0.828 $\pm$ 0.025 & 0.350 $\pm$ 0.035 & 0.323 $\pm$ 0.000 & 0.479 $\pm$ 0.019 & 0.121 $\pm$ 0.060  \\
        \midrule
        {COSINE}    & 0.874 $\pm$ 0.003 & 0.852 $\pm$ 0.013 & 0.836 $\pm$ 0.015 & 0.825 $\pm$ 0.012  & 0.884 $\pm$ 0.048 & 0.323 $\pm$ 0.000 & 0.804 $\pm$ 0.033 & 0.329 $\pm$ 0.007 & 0.619 $\pm$ 0.024 & 0.385 $\pm$ 0.002 \\
        {Denoise}   & 0.868 $\pm$ 0.003 & 0.202 $\pm$ 0.002 & 0.363 $\pm$ 0.243 & 0.246 $\pm$ 0.096  & \textbf{0.954 $\pm$ 0.002} & 0.559 $\pm$ 0.037 & 0.323 $\pm$ 0.000 & 0.323 $\pm$ 0.000 & 0.473 $\pm$ 0.015 & 0.066 $\pm$ 0.015 \\
        \midrule
        \midrule
        {ARS2}  & \textbf{0.893 $\pm$ 0.007} & \textbf{0.890 $\pm$ 0.011} & \textbf{0.868 $\pm$ 0.026} & \textbf{0.851 $\pm$ 0.024}          & \textbf{0.954 $\pm$ 0.004} & \textbf{0.956 $\pm$ 0.005} & \textbf{0.955 $\pm$ 0.007} & \textbf{0.910 $\pm$ 0.048} & \textbf{0.647 $\pm$ 0.022} & \textbf{0.500 $\pm$ 0.008} \\
    \bottomrule
    \end{tabular}
    }
    \caption{\textbf{F1-macro result on RoBERTa}. Comparison among imbalance learning methods, weak supervision methods, and ASR2 (as well as its variants). Note that ARS2 outperform all baselines in all imbalanced datasets.}
    \label{tab:robertares}
\end{table*}
\subsection{Quality Analysis of Sampled Data}
Figure~\ref{fig:agnewsabla} and Figure~\ref{fig:yelpabla} illustrate the cleanliness of the selected data from AGNews and Yelp at each step. 
It may be observed that ARS2 performed similarly to ARS2 (w/o RR) on the balanced set, which means that combining CR and RR can address the problem of decreasing cleanliness caused by RR. 
The results also shows that there was noise in the judgment of rules, and using RR alone could not eliminate the weak labels with noise; rather, it further increased the noise of the data. 
ARS2 generally outperforms a single method with increasing values of the imbalance ratio because the model learns the features of clean data and uses these features to filter out more clean data, gradually improving the cleanliness of the dataset and improving training performance.

\subsection{Performance on Fine-tuning Pre-trained Model}
From Table~\ref{tab:robertares} it may be observed that our optimized RoBERTa training process was able to achieve state-of-the-art results. 
On the Yelp dataset with $\rho=50$, all the baselines could not be fitted well, but the data filtered by our method enable RoBERTa to fit well even under such extreme conditions. 
This occurred because the fine-tuning process of RoBERTa only requires a small amount of data. Hence, we used warmed-up MLP to dynamically choose a small amount of clean and balanced data for training according to the batch size of RoBERTa to achieve good results.
When training RoBERTa, we also linearly increased the training set to expose the model to a greater diversity of data.

\section{Related Works}
\paragraph{Weak Supervision.} 
Weak supervision aims to reduce the cost of annotation, and has been widely applied to perform both classification~\cite{Ratner16, Ratner19, fu2020fast, cosine, denoise} and sequence tagging~\cite{lison2020named, nguyen2017aggregating, safranchik2020weakly, li2021bertifying, lan2020connet} to help reduce human labor required for annotation. Weak supervision builds on many previous approaches in machine learning, such as distant supervision \cite{mintz2009distant, Hoffmann2011KnowledgeBasedWS, Takamatsu2012ReducingWL}, crowdsourcing \cite{Gao2011HarnessingTC, Krishna2016VisualGC}, co-training methods \cite{Blum1998CombiningLA}, pattern-based supervision \cite{Gupta2014ImprovedPL}, and feature annotation \cite{Mann2010GeneralizedEC, Zaidan2008ModelingAA}.  
Specifically, weak supervision methods take multiple noisy supervision sources and an unlabeled dataset as input, aiming to generate training labels to train an end model (two-stage method) or directly produce the end model for the downstream task (single stage method) without any manual annotation. 
Interested readers are referred to a recent survey~\cite{zhang2022survey} for a brief review of the literature on weak supervision.

\paragraph{Class Imbalance Learning.}
Four primary methods has been proposed to solve data imbalance problem, including post-hoc correction, loss weighting, data modification and margin modification. 
\textbf{Post-hoc correction} modifies the logit computed by the model by using the prior of data to bias the training process towards fewer classes with fewer data~\cite{fawcett1996combining, provost2000machine, maloof2003learning, king2001logistic, collell2016reviving, kim2020adjusting, kang2019decoupling}.  
\textbf{Loss weighting} weights the loss by the prior distribution of the training set, so that the data of the minority class exhibits a larger loss, and thus the model learns more bias towards the minority class to balance the minority and majority classes~\cite{xie1989logit, morik1999combining, menon2013statistical, effective_number, fan2017learning}. 
\textbf{Data modification} balances a training set by increasing the number of data samples of minority classes or decreasing the data for majority classes so that the model is trained without favoring the majority classes or ignoring the minority classes~\cite{kubat1997addressing, wallace2011class, chawla2002smote}.
\textbf{Margin modification} balances the minority and majority classes by increasing the margin of minority class data and majority class data, which makes it easier for a model to learn a discriminative, robust decision boundary between the minority and majority class data~\cite{masnadi2010risk, iranmehr2019cost, zhang2017range, cao2019learning, tan2020equalization}. 
In the case of training on a noisy training set, adding weights or adding data may give  incorrect results, causing the model to learn to be biased towards noise.

\section{Conclusion}
In this article, we have proposed ARS2 as a method based on the WS paradigm to reduce both noise and the impact of natural imbalances in data. 
We have proposed PMS to evaluate the level of noise in training data. 
To reduce the impact of noisy data on training, we have proposed two ranking strategies based on PMS, including CR and RR.
Finally, adaptive sampling is performed on the data based on this ranking to clean the data.
We have also presented experimental results on eight different datasets, which demonstrate that ARS2 outperformed traditional WS and loss modification methods.

\section{Limitation} First, because the presented work is focused on adaptive clean data sampling, we use the MLP as a teacher model for a large language model like RoBERTa. 
In the future research, we can consider using co-teaching methods~\cite{han2018co}, which provide a more efficient teacher-student structure to train large models, to further improve the efficiency of teacher model. 
Also, due to the lack of computational resources, we only used a 2-layer MLP and RoBERTa as our backbone model. 
A larger language model like RoBERTa-large~\cite{liu2019roberta} could be considered in future work. 
Finally, the proposed approach could be extended to other tasks such as sequence labeling or natural language inference.

\bibliographystyle{acl_natbib}
\bibliography{references}

\clearpage
\appendix
\section{Datasets Details}
\subsection{Data Source}
\label{appendix:data}
We use the data from WRENCH benchmark~\cite{zhang2021wrench}. 

\paragraph{AGNews}: Dataset is available at \url{https://drive.google.com/drive/u/1/folders/1VFJeVCvckD5-qAd5Sdln4k4zJoryiEun}.

\paragraph{Yelp}: The raw dataset is available at \url{https://huggingface.co/datasets/yelp_review_full}. The preprocessed dataset is available at \url{https://drive.google.com/drive/u/1/folders/1VFJeVCvckD5-qAd5Sdln4k4zJoryiEun}.

\paragraph{TREC, Chemprot}: The preprocessed dataset is available at \url{https://drive.google.com/drive/u/1/folders/1VFJeVCvckD5-qAd5Sdln4k4zJoryiEun}.

\begin{figure*}[!t]
    \centering
    \subfigure[AGNews]{
        \includegraphics[width=\textwidth]{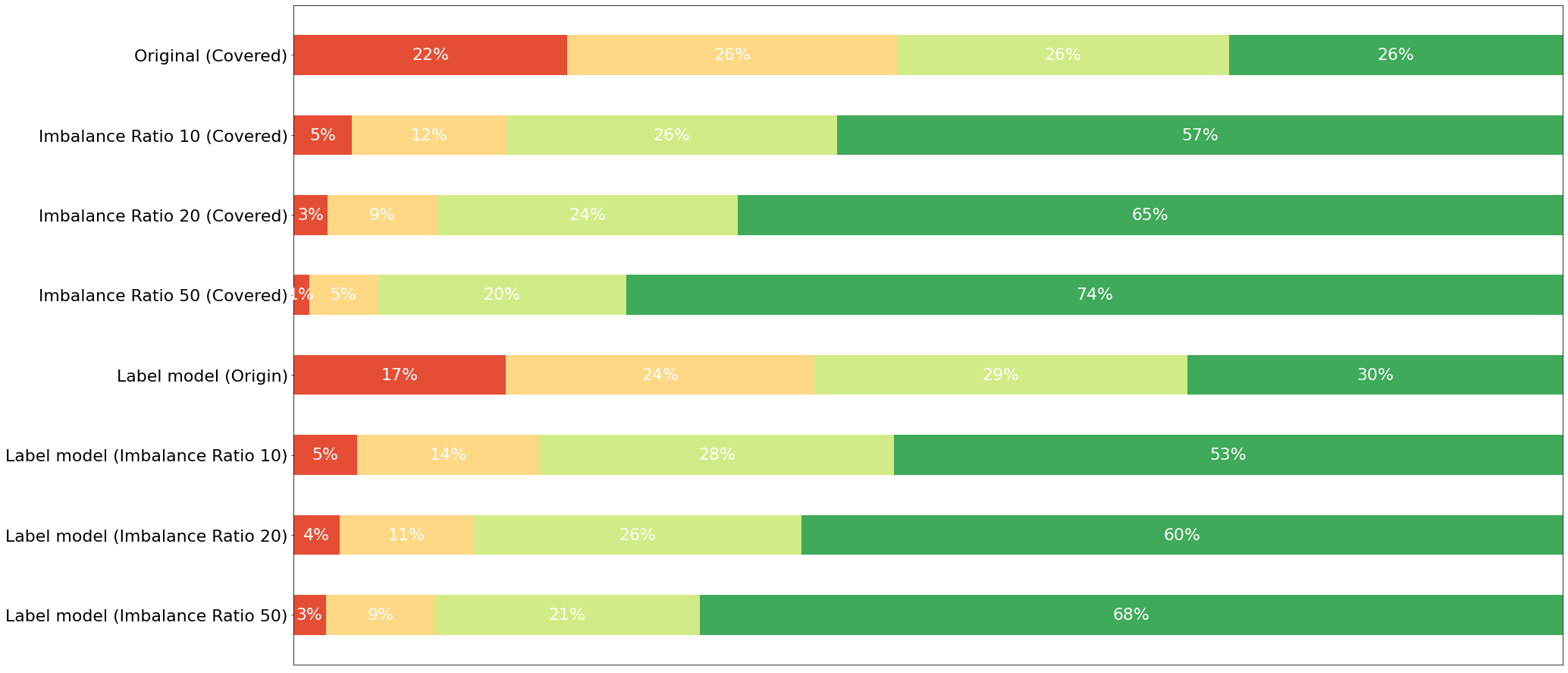}
    }
    \subfigure[Yelp]{
        \includegraphics[width=\textwidth]{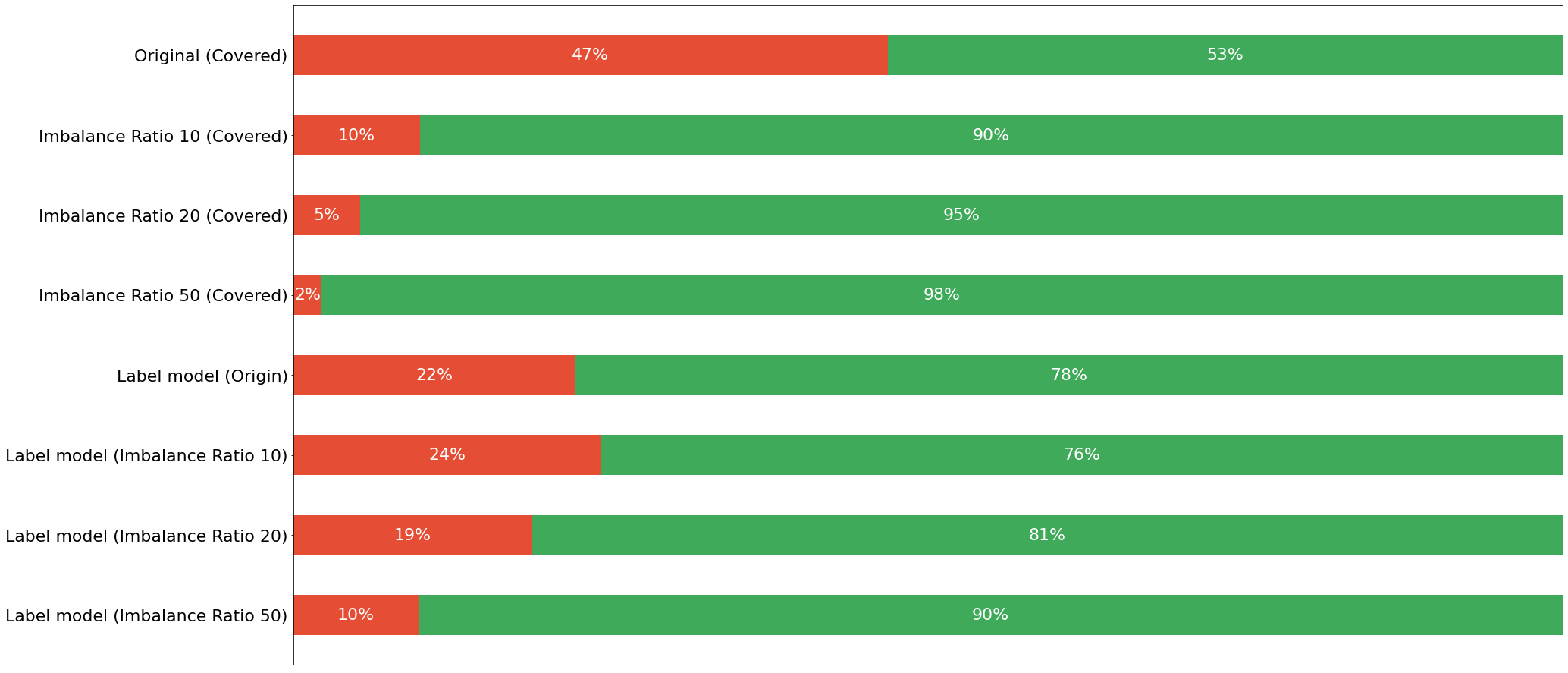}
    }
    \subfigure[Chemprot]{
        \includegraphics[width=\textwidth]{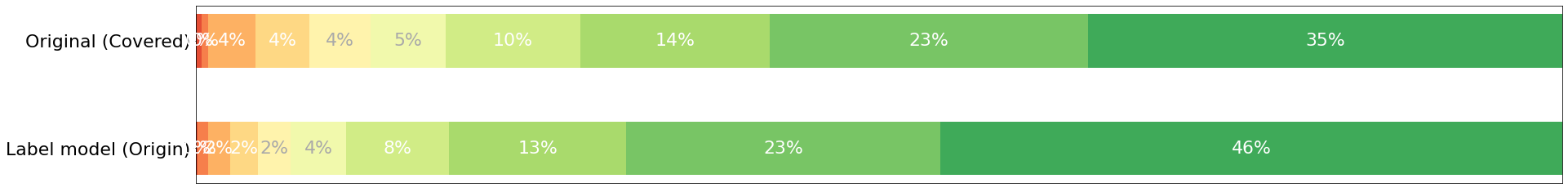}
    }
    \subfigure[TREC]{
        \includegraphics[width=\textwidth]{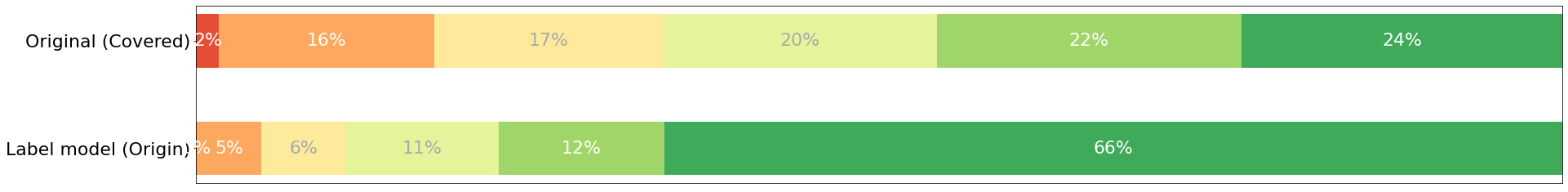}
    }
    \caption{Label statistic of each dataset.}
    \label{fig:labelstat}
\end{figure*}

For these two datasets, we design eight different imbalance ratios to evaluate all methods. The details are shown in Table~\ref{tab:imbalance} and the label distribution for each dataset are shown in Fig~\ref{fig:labelstat}.
\begin{table}[!htb]
    \begin{adjustbox}{max width=0.48\textwidth}
    \begin{tabular}{ l c c c c c }
    \toprule
        \textbf{Dataset ($\downarrow$)} & \textbf{Imbalance Ratio $\rho$ ($\downarrow$)} & \textbf{Training Noise} & \textbf{\#Train} & \textbf{\#Train (Covered)} & \textbf{\#Valid} \\
    \midrule
    \multirow{4}{*}{AGNews}    
        & {1}  & 18.4\% & 96.0k & 66.3k & 12.0k \\
        & {10} & 12.8\% & 42.7k & 30.5k & 5.4k \\ 
        & {20} & 12.9\% & 37.3k & 26.9k & 4.7k \\
        & {50} & 12.5\% & 32.8k & 23.7k & 4.1k \\
    \midrule[0.05pt]
    
    \multirow{4}{*}{Yelp}    
        & {1}  & 29.8\% & 30.4k & 25.2k & 3.8k \\
        & {10} & 22.3\% & 16.8k & 13.2k & 2.0k \\
        & {20} & 18.1\% & 16.0k & 12.5k & 1.9k \\
        & {50} & 10.2\% & 15.6k & 12.1k & 1.9k \\
    \bottomrule
    \end{tabular}
    \end{adjustbox}
    \caption{Weakly Annotated Imbalance Dataset Statistics.}
    \label{tab:imbalance}
\end{table}

\section{Details on Implementation and Experiment Setups}
\label{appendix:exp}

\subsection{Computing Infrastructure}
\textit{System}: Windows Subsystem Linux 2; Python 3.8; PyTorch 1.9. \\
\textit{CPU}: Intel(R) Core(TM) i9-12900K CPU. \\
\textit{GPU}: NVIDIA RTX 3090. \\

\begin{table*}[t]
    \centering
    \scalebox{0.73}{
    \begin{tabular}{ l c c c c c c c c c }
    \toprule
         \multicolumn{1}{c}{} &
         \multicolumn{1}{c}{} &
         \multicolumn{4}{c}{\textbf{Agnews (Imbalance Ratio ($\downarrow$))}} &
         \multicolumn{4}{c}{\textbf{Yelp (Imbalance Ratio ($\downarrow$))}}   \\
    \midrule
    \multirow{1}{0pt}{}
        \textbf{Methods} & \textbf{Hyper-parameter} & 1 & 10 & 20 & 50 & 1 & 10 & 20 & 50\\
    \midrule
    \multirow{1}{0pt}{} 
        & Batch Size & \multicolumn{8}{@{\hskip1pt}c@{\hskip1pt}}{128} \\
        & $\gamma$ & 3 & 10 & 1 & 1 & 1 & 1 & 1 & 1 \\
        & $\xi$ & -0.2 & -0.3 & 0 & -0.2 & 0 & -0.1 & -0.1 & -0.3 \\
    \midrule
    \multirow{2}{0pt}{CE+LA}
        & Dropout Ratio & 0.2 & 0.2 & 0.0 & 0.2 & 0.2 & 0.0 & 0.0 & 0.0 \\
        & Learning Rate & 0.001 & 0.001 & 0.003 & 0.001 & 0.001 & 0.0001 & 0.0001 & 0.0006 \\
    \midrule
    \multirow{4}{0pt}{CE+EN}
        & Dropout Ratio & 0.2 & 0.0 & 0.0 & 0.0 & 0.2 & 0.0 & 0.0 & 0.0 \\
        & Learning Rate & 0.001 & 0.001 & 0.001 & 0.001 & 0.001 & 0.001 & 0.001 & 0.001 \\
        & $\beta_{EN}$ & 0.9999 & 0.999 & 0.9 & 0.99 & 0.9999 & 0.99 & 0.999 & 0.9999 \\
        & $\gamma_{EN}$ & 2.0 & 0.5 & 1.0 & 2.0 & 1.0 & 0.5 & 1.0 & 0.5 \\
    \midrule
    \multirow{4}{0pt}{CE+EN+LA}
        & Dropout Ratio & 0.0 & 0.2 & 0.2 & 0.2 & 0.2 & 0.0 & 0.0 & 0.0 \\
        & Learning Rate & 0.0001 & 0.001 & 0.001 & 0.001 & 0.001 & 0.01 & 0.001 & 0.001 \\
        & $\beta_{EN}$ & 0.9999 & 0.99 & 0.999 & 0.99 & 0.999 & 0.9 & 0.999 & 0.99 \\
        & $\gamma_{EN}$ & 2.0 & 0.5 & 0.5 & 0.5 & 2.0 & 1.0 & 0.5 & 0.5 \\
    \midrule
    \multirow{5}{0pt}{Dice}
        & Dropout Ratio & 0.0 & 0.2 & 0.0 & 0.2 & 0.0 & 0.0 & 0.2 & 0.0 \\
        & Learning Rate & 0.001 & 0.001 & 0.001 & 0.001 & 0.01 & 0.01 & 0.001 & 0.001 \\
        & $\alpha$ & 0.2 & 0.2 & 0.2 & 0.2 & 0.1 & 0.1 & 0.1 & 0.1 \\
        & $\gamma_{Dice}$ & 0.005 & 1.0 & 0.268 & 0.005 & 1.0 & 0.268 & 0.001 & 0.001 \\
        & Denominator Square & False & True & True & False & True & True & False & False \\
    \midrule
    \multirow{4}{0pt}{LDAM}
        & Dropout Ratio & 0.2 & 0.0 & 0.2 & 0.0 & 0.2 & 0.0 & 0.0 & 0.0 \\
        & Learning Rate & 0.001 & 0.001 & 0.001 & 1e-5 & 0.001 & 0.001 & 1e-5 & 1e-5 \\
        & max margin & 0.7 & 0.8 & 0.9 & 0.2 & 0.8 & 0.6 & 0.6 & 0.3 \\
        & $s$ & 4.0 & 1.0 & 1.0 & 16.0 & 4.0 & 1.0 & 1.0 & 10.0 \\
    \midrule
    \multirow{4}{0pt}{COSINE}
        & Learning Rate & 1e-5 & 1e-5 & 3e-5 & 1e-5 & 3e-5 & 1e-5 & 3e-5 & 3e-5 \\
        & $T_3$ & 100 & 50 & 50 & 200 & 200 &  100 & 50 & 200 \\
        & $\lambda$ & 0.1 & 0.1 & 0.1 & 0.1 &  0.1 & 0.01 & 0.05 & 0.1 \\
        & $\gamma_C$ & 0.3 & 0.9 & 0.9 & 0.7 &  0.3 & 0.9 & 0.9 & 0.3 \\
    \midrule
    \multirow{5}{0pt}{Denoise}
        & Learning Rate & 1e-5 & 3e-6 & 1e-6 & 3e-6 & 3e-5 & 3e-5 & 1e-5 & 3e-6 \\
        & Hidden Size & 512 & 512 & 256 & 512 & 128 & 64 & 256 & 256 \\
        & $\alpha$ & \multicolumn{8}{@{\hskip1pt}c@{\hskip1pt}}{0.6} \\
        & $c_1$ & 0.1 & 0.1 & 0.3 & 0.3 &  0.7 & 0.3 & 0.5 & 0.1 \\
        & $c_2$ & 0.7 & 0.7 & 0.3 & 0.3 &  0.1 & 0.5 & 0.1 & 0.5 \\
    \bottomrule
    \end{tabular}
    }
    \caption{Hyperparameter configurations for all baselines.}
    \label{tab:hyperparameter}
\end{table*}

\subsection{Number of Parameters}
ARS2 and all baselines use 2-layer MLP and Roberta-base~\cite{liu2019roberta} with a task-specific classification head on the top as the backbone. The 2-layer MLP has 128 neural unit in each layer, the total parameters are $128\times128\times number \ of \ classes$. The Roberta-base model contains 125M trainable parameters, and we fine-tune the last 4 layers with token size 512. We do not introduce any other parameters in our experiments.

\subsection{Experiment Setups}
\label{apdx:setups}
All of our methods and baselines are run with 5 different random seeds and the result is based on the average performance on them. This indeed creates $8$ (the number of datasets with four imbalance ratios) $\times$ $5$ (the number of random seeds) $\times$ $12$ (the number of methods) $\times$ $2$ (the number of end models, MLP and RoBERTa) = $960$ experiments for fine-tuning, which is almost the limit of our computational resources, not to mention grid search for hyperparameter for each method.
We have shown both the mean and the standard deviation of the performance criteria in our experiment sections.

\subsection{Implementations Baselines}
For these three methods listed below, since they are mainly used in CV tasks; thus the code is hard to directly used for our experiments. We re-implement these methods based on their implementations in WRENCH codebase.

\paragraph{EN}: \url{https://github.com/richardaecn/class-balanced-loss}.

\paragraph{LA}:  \url{https://github.com/google-research/google-research/tree/master/logit_adjustment}.

\paragraph{LDAM}:  \url{https://github.com/kaidic/LDAM-DRW}.

For these two weak supervision baselines listed below, we use the implementation provided by WRENCH.

\paragraph{COSINE, Denoise}:  \url{https://github.com/JieyuZ2/wrench}.

Our implementation of ARS2 will be published upon acceptance.

\subsection{Hyperparameters for General Experiments}
\label{appendix:hyper}
We use AdamW \cite{loshchilov2018decoupled} as the optimizer, and the learning rate of 2-layer MLP is chosen from $1\times10^{-5}$ to $1\times10^{-1}$, and $1\times10^{-5}, 3\times10^{-5}, 1\times10^{-6} ,3\times10^{-6}$ for RoBERTa. Dropout rate of 2-layer MLP is chosen from $0, 0.2$. We set weight decay rate as $0$ and batch size as $128$ for 2-layer MLP, and weight decay rate as $1\times10^{-4}$ and batch size as $16$ for RoBERTa. We warm-up and continual training until early stop, and evaluate 2-layer MLP in every 100 steps and RoBERTa every 5 steps. Finally, we use the model with the best performance on the development set for testing. The general hyperparameters we use are shown in Table~\ref{tab:hyperparameter}.

\subsection{Hyperparameters for ARS2}
\label{appendix:ours}
The hyperparameter of ARS2 includes $\gamma, \xi$ thus it does not require heavy hyperparameter tuning. In our experiments, we search $\gamma$ from $1$ to $10$, and $\xi$ from $-0.5$ to $0.5$.

\subsection{Hyperparameters for Loss Modification Baselines}
For loss modification methods, we mainly tune their key hyperparameters. For EN~\cite{effective_number}, we tune the number for effective number $\beta$ from $[0.9, 0.99, 0.999, 0.9999]$, $\gamma$ from $[0.5, 1, 2]$ and report the best performance. For LA~\cite{logit_adjustment}, they use $\tau$ to scale the prior ratio. But in our experiment, we directly set $\tau$ as $1$ to achieve LA's performance as efficiently as possible. For Dice~\cite{li2019dice}, it use $\alpha$ to scale the $(1-p)$ factor avoiding it become too large, we search $\alpha$ from $0.1$ to $1$. $\gamma_{Dice}$ is designed to help numerator and denominator become smooth, we search $\gamma_{Dice}$ from $1\times10^{-4}$ to 1. Denominator square is designed to control the size of denominator, we search this hyperparameter in [True, False]. For LDAM~\cite{ldam}, we search max margin from $0.1$ to $0.9$ and $s$ from $1$ to $30$.

\subsection{Hyperparameters for Weak Supervision Baselines}
For COSINE~\cite{cosine}, we search learning rate from $[1\times10^{-5}, 3\times10^{-5}, 1\times10^{-6}, 3\times10^{-6}]$, teacher model update frequency from $[50, 100, 200]$, regularize power scale from $[0.01, 0.05, 0.1]$ and margin threshold $\gamma_C$ from $[0.1, 0.3, 0.5, 0.7, 0.9]$. For Denoise~\cite{denoise}, we search learning rate from $[1\times10^{-5}, 3\times10^{-5}, 1\times10^{-6}, 3\times10^{-6}]$, denoiser hidden size from $[64, 128, 256, 512]$ and $c_1, c_2$ from $[0.1, 0.3, 0.5, 0.7, 0.9]$.

\end{document}